\documentclass[sigconf]{acmart}
\AtBeginDocument{%
  }

\copyrightyear{2026}
\acmYear{2026}
\setcopyright{none}
\setcctype{by}
\acmConference[KDD '26]{Proceedings of the 32nd ACM SIGKDD Conference on Knowledge Discovery and Data Mining V.1}{August 09--13, 2026}{Jeju Island, Republic of Korea}
\acmBooktitle{Proceedings of the 32nd ACM SIGKDD Conference on Knowledge Discovery and Data Mining V.1 (KDD '26), August 09--13, 2026, Jeju Island, Republic of Korea}
\acmPrice{}
\acmDOI{10.1145/3770854.3780305}
\acmISBN{979-8-4007-2258-5/2026/08}

\usepackage{multirow}
\usepackage{subcaption}
\usepackage{algorithmic}
\usepackage{algorithm}

\begin{document}

\title{In-context Learning of Evolving Data Streams with Tabular Foundational Models}

\author{Afonso Lourenço}
\orcid{0000-0002-3465-3419}
\affiliation{%
  \institution{GECAD, ISEP, Polytechnic of Porto}
  \streetaddress{Rua Dr. António Bernardino de Almeida}
  \city{Porto}
  \country{Portugal}
  \postcode{4249-015}
}

\author{João Gama}
\orcid{0000-0003-3357-1195}
\affiliation{%
  \institution{INESC-TEC, FEP, University of Porto}
  \streetaddress{Rua Dr. Roberto Frias}
  \city{Porto}
  \country{Portugal}
  \postcode{4200-465}
}

\author{Eric P. Xing}
\orcid{0000-0002-3683-4280}
\affiliation{%
  \institution{Carnegie Mellon University}
  \institution{Mohamed bin Zayed University of AI}
  \city{Pittsburgh}
  \country{USA}
  \postcode{PA 15213}
}

\author{Goreti Marreiros}
\orcid{0000-0003-4417-8401}
\affiliation{%
  \institution{GECAD, ISEP, Polytechnic of Porto}
  \streetaddress{Rua Dr. António Bernardino de Almeida}
  \city{Porto}
  \country{Portugal}
  \postcode{4249-015}
}

\renewcommand{\shortauthors}{Lourenço et al.}

\begin{abstract}
State-of-the-art data stream mining has long drawn from ensembles of the Very Fast Decision Tree, a seminal algorithm honored with the 2015 KDD Test-of-Time Award. However, the emergence of large tabular models, i.e., transformers designed for structured numerical data, marks a significant paradigm shift. These models move beyond traditional weight updates, instead employing in-context learning through prompt tuning. By using on-the-fly sketches to summarize unbounded streaming data, one can feed this information into a pre-trained model for efficient processing. This work bridges advancements from both areas, highlighting how transformers' implicit meta-learning abilities, pre-training on drifting natural data, and reliance on context optimization directly address the core challenges of adaptive learning in dynamic environments. Exploring real-time model adaptation, this research demonstrates that TabPFN, coupled with a simple sliding memory strategy, consistently outperforms ensembles of Hoeffding trees, such as Adaptive Random Forest, and Streaming Random Patches, across all non-stationary benchmarks.
\end{abstract}




\maketitle

\begin{figure}[htbp]  
    \centering
    \begin{subfigure}{0.24\textwidth}
        \centering
        \includegraphics[width=\textwidth]{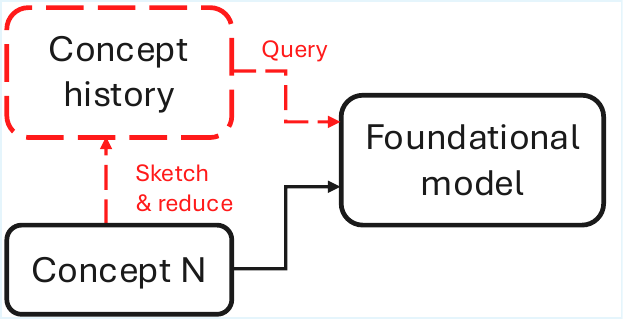}
        \caption{In-context stream mining}
    \end{subfigure}
    \begin{subfigure}{0.23\textwidth}
        \centering
        \includegraphics[width=\textwidth]{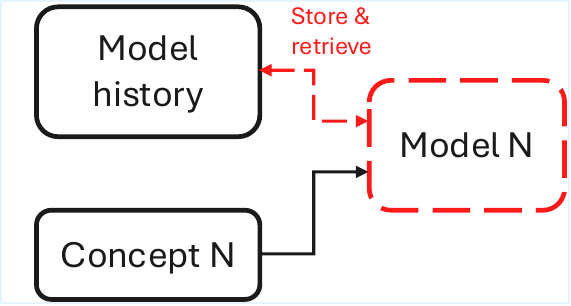}
        \caption{Conventional stream mining}
    \end{subfigure}
    \caption{A new paradigm}
    \label{fig:newparadigm}
\end{figure}

\section{Introduction}

Data stream mining is an area in machine learning where the inference and training of the algorithm are performed in real-time, dealing with large volumes of ever-evolving tabular data \cite{gama2010knowledge}. Differently to batch learning where all training data necessary to induce a model is available, data streams incrementally arrive at any time. Thus, making a model easily outdated due to the occurrence of concept drifts, i.e., distribution changes over time \cite{gama2014survey}. To circumvent these challenges, many stream mining algorithms have been proposed with different requirements being imposed: (1) be ready to predict and update the model at any point, and in limited time; (2) process an example at a time, inspecting only once with a limited sized model; (3) be able to adapt to change. 

\textbf{Decision trees.} In fulfilling these requirements, the current state-of-art for supervised learning in tabular data streams has long been incremental decision trees (IDTs), whose success can be attributed to two key factors: approximation-based splitting techniques \cite{lourenco2025dfdt} and effective adaptation strategies \cite{neves2025online}. Approximation-based splitting involves updating statistical summaries of entropy-based metrics, e.g. information gain, and determining whether the observed utility of a split is statistically close to its true utility when the data distribution is unknown, e.g. via the Hoeffding bound \cite{domingos2001catching}. However, as tree grows from the root node, the descendant nodes subsequently get fixed to covering particular sub-spaces of the space covered by their parent node. To address this, adaptation strategies allow to trigger the forgetting of old knowledge, pruning the affected parent nodes, instead of a complete replacement of the base learner \cite{bifet2009adaptive}. Alternatively, multiple trees can be combined into a high diversity ensemble, which naturally develops decision boundaries that locally adapt to changes in the data distribution \cite{gomes2019streaming,krawczyk2017ensemble}. Upon a concept drift, the diversity among trees can be exploited as some local minima are less affected than others. Affected components can be dynamically deleted, added, and combined with a voting scheme, weighting learners based on their past performance. Moreover, meta-learning \cite{anderson2016recurring} and model-based clustering \cite{xu2016mining} allows to more explicitly model history mechanisms to retain a pool of previous active and inactive concepts.

\textbf{Lack of autonomy.} Regardless of how sophisticated these strategies are, these still rely on human expertise to select the most suitable algorithm, or determine the optimal configuration for a given algorithm. Indeed, combining all these building blocks in a single solution normally yields more inductive biases to assume and hyperparameters to tune. Consequently, as new concepts emerge, the performance curves of different algorithm settings cross. To address this, one can simply build heterogeneous ensembles with different algorithms, whose weights are dynamically selected based on recent performance \cite{van2015having}. Alternatively, one can calculate metafeatures over sliding windows to build a meta-model that predicts the best algorithm \cite{van2014algorithm}, the best drift detector \cite{aguiar2023enhancing}, or the most suitable uncertainty threshold for active learning strategies \cite{martins2023meta}, in the next interval of unseen data.

\textbf{Foundational models (FMs).} Conversely, the disruptive success and knowledge-modeling capabilities of foundational models \cite{brown2020language} brings the opportunity to tap into prior experience, to the point that most of the required learning has already been done beforehand. While the aforementioned streaming methods follow the traditional two-stage methodology, i.e., first optimizing the model with several tuning iterations over a sliding window to then select the best algorithm or hyperparameter configuration for the next window, FMs allow to instantaneously deploy a model, bypassing the computational and temporal costs of this traditional process. By pretraining on extremely large corpora of datasets, FMs develop common sense inductive biases from the infinite amounts of prior learning experiences. Thus, driving emergent abilities that improve with scale, such as few-shot in-context learning (ICL), which allows models to perform new tasks at inference time by conditioning on a set of input-output examples, without any parameter updates \cite{brown2020language}.

\textbf{Textual tabular prompts.} Consequently, this ICL ability pushed a new research paradigm, focused on adapting pre-trained models through prompt tuning rather than adjusting model weights, by discrete search with natural language, or optimizing the model's embedding space directly. Within this paradigm and the emerging possibilities of multimodal applications, various efforts were made for flexible textual prompts of structured tabular data with heterogeneous feature spaces and mixed-type columns \cite{fang2024large}. As language-based FMs' capabilities on table understanding were validated, new efforts went towards directly applying them for supervised learning of tabular data, i.e. predicting an unseen sample by continuing its textual description on the target column \cite{dinh2022lift, hegselmann2023tabllm}. However, despite the potential of these advancements to perform semantically-driven drift detection and model adaptation, several limitations are expected to persist in the near term hindering their application for data stream mining: limited context windows, quadratic complexity costs, struggle with autoregressively understanding continuous variables \cite{thawani2021representing,van2024latable}, and sensitivity to unexpected characters \cite{zhao2021calibrate}.

\textbf{Large Tabular Models (LTMs).} To alleviate all these limitations, one can shift to designing a dedicated architecture for numeric data from scratch, completely pre-trained on a wide range of tabular datasets with different data distributions, allowing it to learn relevant and general meta-features. These are referred to as LTMs \cite{van2024tabular}. While it has been known for a while that transformers can handle variable length input sequences in order to perform pretraining and cross-table knowledge transfer \cite{wang2022transtab}, it has been recently demonstrated that finetuning from a pretrained tabular transformer \cite{gorishniy2021revisiting, wu2021fastformer, somepalli2021saint} is superior to training tabular transformers from scratch \cite{zhu2023xtab}. Following these findings, various LTMs pretrained on synthetic tabular data have been proposed for classification \cite{hollmann2025accurate,bonet2024hyperfast,ye2023training}. For instance, the original TabPFN \cite{hollmann2022tabpfn} is pre-trained with a 12-layer Transformer for 18000 batches of 512 synthetically generated datasets generated using randomly initialized neural networks to impose the diverse interrelations that exist among the features of realistic tabular datasets, e.g. causal relationships and feature importance  \cite{muller2021transformers}. While this training step is moderately expensive, requiring a total of 20 hours on 8 GPUs (Nvidia RTX 2080 Ti), it is done offline, in advance, and only once, as part of the algorithm development. Then, in deployment TabPFN performs instant classification without fine-tuning, adapting to unseen datasets in a single forward pass at inference time by using various training examples as context, analogously to how LLMs use the preceding tokens \cite{brown2020language}. However, in contrast to LLMs, LTMs like TabPFN allow scaling to much larger datasets, saving tuning resources during deployment, while providing consistent behavior supported by traditional statistical learning.

\textbf{This work.} Building on these advancements, this work proposes a new paradigm, using on-the-fly techniques to summarize unbounded data streams before feeding them to LTMs. While typical streaming methods rely on adapting the model parameters to deal with non-stationary data distributions, this new paradigm uses input data as learnable parameters without changing model parameters, decoupling the design of the learning algorithm from the foundational model. The main idea is to transform big data into a sketch whose size and running time has little or no dependency on the size of the data stream, as illustrated in Figure \ref{fig:newparadigm}. The learning algorithm is then applied to this dynamically contextualized sketch, with the main challenge of balancing the trade-off between sketch size and information loss. Thus, shifting from in-weights learning to in-context learning. The main contributions are outlined as follows:

\begin{itemize}
    \item Introduce a novel in-context learning paradigm for tabular data streams, formalizing how the distributional qualities of the pre-training data and a mesa-optimization architectural bias address the core challenges of adaptive learning. 
    \item Extend the popular TabPFN transformer \cite{hollmann2022tabpfn} with a new inference-time sketching mechanism, without altering its pretrained weights. Inspired by a bayesian and frequentistic perspective, we introduce a bias-variance decomposition of prediction error, proving variance reduction through more in-context examples, and addressing drifting context-induced biases via a dual-memory FIFO mechanism that maintains inter-class balance and captures intra-class variability through long- and short-term buffers.
\end{itemize}

\section{In-context stream learning}
\label{sec:methodology}

ICL requires a model to implicitly construct a map from in-context examples to a predictor of a new concept on which it has not been previously trained, without any updates to the model's parameters themselves. Indeed, it has been show empirically that transformers with fixed parameters can learn linear functions comparable to the optimal least squares estimator, by encoding smaller models in their activations, and updating these implicit models as new examples appear in the context on the fly \cite{von2022what}. Furthermore, it has been demonstrated that transformers implement more complex function classes implicitly, i.e., sparse linear functions, two-layer neural networks, and decision trees, with performance that matches or exceeds task-specific learning algorithms \cite{min2022rethinking}. However, it remains unclear to what extent these models are able to learn new tasks from in-context examples alone as opposed to indexing into a vast set of known tasks from the training data. In this regard, it has been hypothesized that this emergent behavior of transformers is driven by both the distributional qualities of the training data itself \cite{chan2022data} and an architectural bias where the internal structure of the model ends up performing mesa-optimization \cite{bai2022uncovering}.

\textbf{Natural data hypothesis.} The data hypothesis was inspired by the observation that many natural data sources differ from typical batch supervised datasets due to a few notable features: (1) natural data is temporally bursty, i.e. a given concept may have a distribution that is not uniform across time, instead tending to appear in clusters \cite{altmann2009beyond}; (2) the meaning of these concepts in natural data is often dynamic rather than fixed, usually in a context-dependent way; (3) the marginal distribution across entities is highly skewed, with large numbers of rarely occurring classes \cite{chan2022data}. Indeed, these are the properties of data streams that motivate the development of incremental learning algorithms. While batch i.i.d. training typically consists of item classes that reoccur with uniform regularity, and input-output mappings that are fixed throughout training, data stream mining implies training on sequences of data with drifting concepts, akin to having many labels per item \cite{read2023multi}, a few highly reoccurring concepts \cite{anderson2016recurring}, and a large number of outliers that reoccur much more rarely, thus excluded from the concept forming process due to their disproportionately less likelihood to occur multiple times within a given context window \cite{de2016minas}. 

\begin{figure}[ht]
    \centering
    \includegraphics[width=0.48\textwidth]{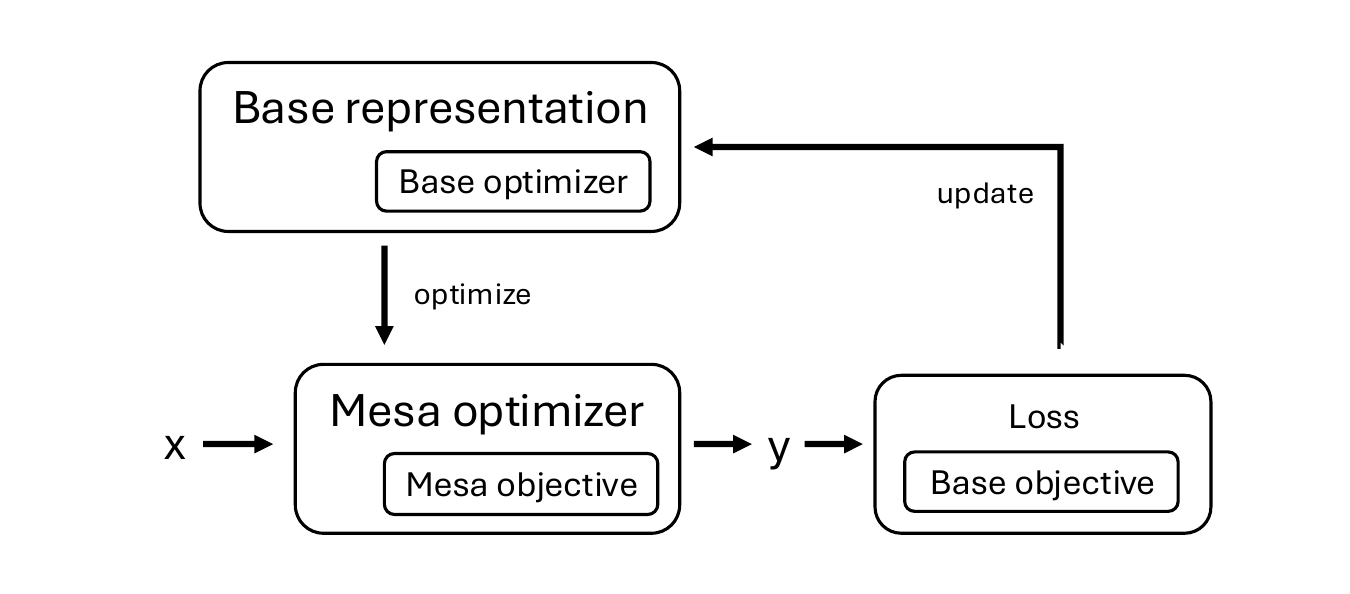}
    \caption{ICL as a mesa-optimization capability}
    \label{fig:mesa}
\end{figure}

\textbf{Mesa-optimization hypothesis.} The algorithm hypothesis identifies this emergent behavior as stemming from an architectural bias where the internal structure of transformers ends up performing mesa-optimization \cite{bai2022uncovering}, which doesn't happen for other recurrent models, like LSTMs and RNNs, even if matched on number of parameters \cite{chan2022data}. Mesa-optimization occurs when a base optimizer is searching for algorithms to solve some problem and finds a model that is itself a mesa-optimizer \cite{hubinger2019risks}. Unlike the base objective, the mesa-objective is not specified directly by the programmers. Instead, a learned process running within the forward pass constructs an internal mesa-objective, and its corresponding solution through mesa-optimization, because it is instrumentally useful for solving the given task. While meta means to go above, thus stepping outside to a higher logical level, mesa means going to the other direction, to go down into the content narrowing the details into smaller and smaller pieces \cite{cheal2021what}. Many reasons have been identified for this phenomenon. Firstly, the base optimizer's incentive to find a compressed policy creates a bias towards simple solutions with low description length. Moreover, the transformer's statefulness favors mesa-optimization by decreasing the implicit penalty on time complexity imposed by enforcing all computations to end when the learned algorithm produces its output. And, the ability of the learned algorithm to save and refer back to previous state enables the caching of intermediate results. These factors, when combined with heavy pre-training on diverse tasks that are likely to be completely novel, incentivize the base optimizer to expend optimization power in learning how to adjust to a new task immediately in order to perform consistently in an diverse environment. Figure \ref{fig:mesa} illustrates this ability.

\textbf{Implicit meta-learning.} While breaking things down into bits is a mesa process, the model always returns to rebuild and re-map the actual details. In going mesa, all generalizations are deconstructed which allows to reconstruct the input-output mapping in a more resourceful manner. In this regard, training a model to perform ICL can also be viewed as an instance of the learning-to-learn or meta-learning paradigm, with learning that takes place on various time scales, fast and slow, comprising both mesa- and meta-processes.
Examples include approximating an implicit model selection procedure competitive with empirical risk minimization \cite{agrawal2022transformers}, a gradient-based few-shot learning within its forward pass \cite{von2022transformers} or a kernel-based sequence model \cite{mccoy2022recasting}. However, in these cases, researchers explicitly designed the meta-training regime to incentivize ICL, whereas in the case of transformers, the capacity for ICL is emergent. Neither the model’s transformer architecture nor its learning objective are explicitly designed with ICL in mind.

\section{Large tabular model}

\begin{figure}[h]
    \centering
    \includegraphics[width=0.5\textwidth]{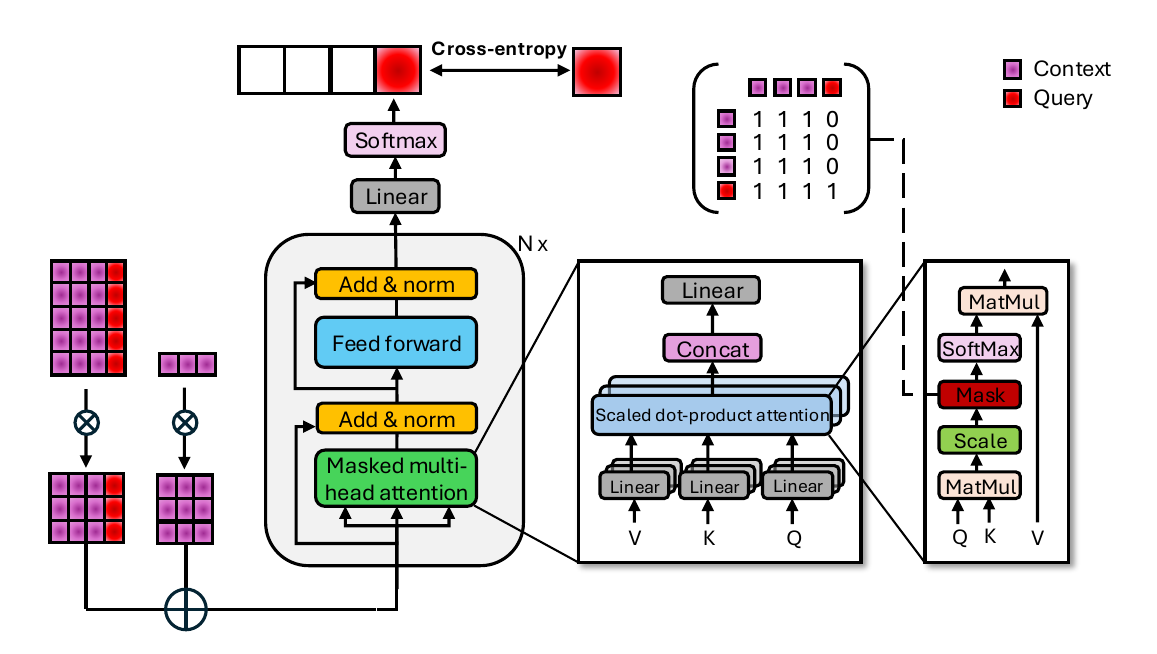}
    \caption{Architecture and training procedure of the foundational model, where $\otimes$ represents features and float targets being separately encoded using a linear layer, and the embedding is pushed through a decoder-only transformer model}
    \label{fig:architecture}
\end{figure}

Since the pre-training stage does not represent the added value of this work, we rely on the popular TabPFN transformer-based model \cite{hollmann2022tabpfn,hollmann2025accurate}, relying on its original dataset generation and architecture, only performing minor modifications on the inference time procedure. At the core of TabPFN is the self-attention mechanism, interchanging information among different positions within a sequence \cite{vaswani2017attention}. In self-attention, each element in the input sequence of length $T$ is linearly projected to produce $d$-dimensional queries, keys, and values, each represented by $Q, K, V \in \mathbb{R}^{T \times d}$. The attention matrix is computed as $\text{softmax}(QK^T/\sqrt{d})$ where softmax is applied row-wise. It is then multiplied to the value matrix to produce the output of the layer $O = \text{softmax}(QK^T/\sqrt{d})V$. If self-attention is computed in this manner, it is more precisely called bi-directional self-attention since every token can attend to all the other tokens in both directions. However, TabPFN relies on causal attention layers typical of a standard decoder-only Transformer architecture \cite{liu2018generating}, making it possible to process a streaming sequence where the tokens become available one at a time \cite{mccoy2022recasting}. The complete architecture is given in Figure \ref{fig:architecture}.

\begin{figure}[ht]
    \centering
    \includegraphics[width=0.48\textwidth]{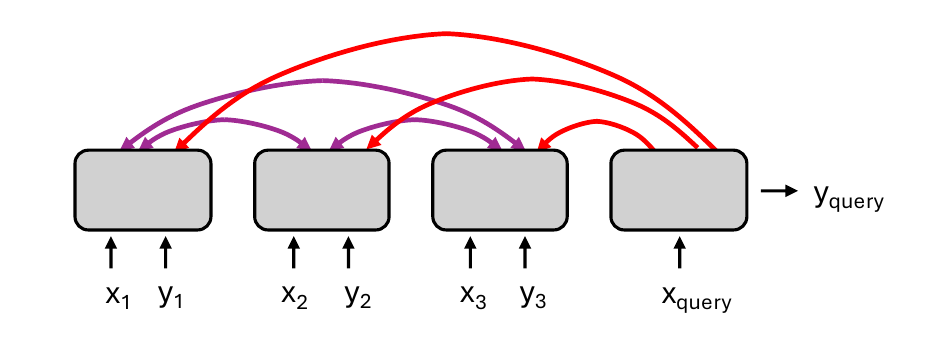}
    \caption{Special attention mask}
    \label{fig:attention}
\end{figure}

\textbf{Architecture.} Each instance's feature vector is treated as a set-valued input to exploit permutation invariance and represented as a token with dimension $d_\text{token}$, allowing token representations to attend to each other for a fixed number of features $d_f$. Firstly, the creation of in-context tokens $H \in \mathbb{R}^{|S| \times d_\text{token}}$ is perfomed by embedding the features and classes linearly using weights $W_x \in \mathbb{R}^{d_f \times d_\text{token}}$ and $w_y \in \mathbb{R}^{d_\text{token}}$ respectively. Then, the embedded tokens are pushed through the standard decoder-only Transformer \cite{liu2018generating}. Each layer in the transformer autoregressively maps the matrix $H^l$ to a sequence $H^{l+1}$, processing each collumn $h_i^l$ of $H^l$ in parallel. Each layer first computes a self attention and then applies a feedforward transformation. However, with a special attention mask, where in-context tokens only attend to each other, with no attention to the single query instance, as illustrated in Figure \ref{fig:attention}. Instead, the target to the held-out query instance is used as a label, minimizing the cross-entropy loss, which approximates the true Bayesian posterior predictive distribution \cite{muller2021transformers}. This procedure results in a computational cost of a forward pass scaling as $O(|S|^2)$. 

\textbf{Learning.} To prior-fit the network as part of the algorithm development, TabPFN repeatedly samples synthetic realistic datasets. These are generated by carefully crafting correlations blockwise, imposing causal relationships between features, and tuning the magnitude of weights for variety in the feature importance, using Bayesian Neural Networks and Structural Causal Models with random characteristics and randomly initialized weights \cite{hollmann2022tabpfn, von2022what}. In this regard, to better understand and quantify the uncertainty in the process of learning a LTM, one can take both a bayesian \cite{xie2022explanation,garg2022transformers} and a frequentistic perspective \cite{nagler2023statistical}.

\begin{figure}[ht]
    \centering
    \includegraphics[width=0.4\textwidth]{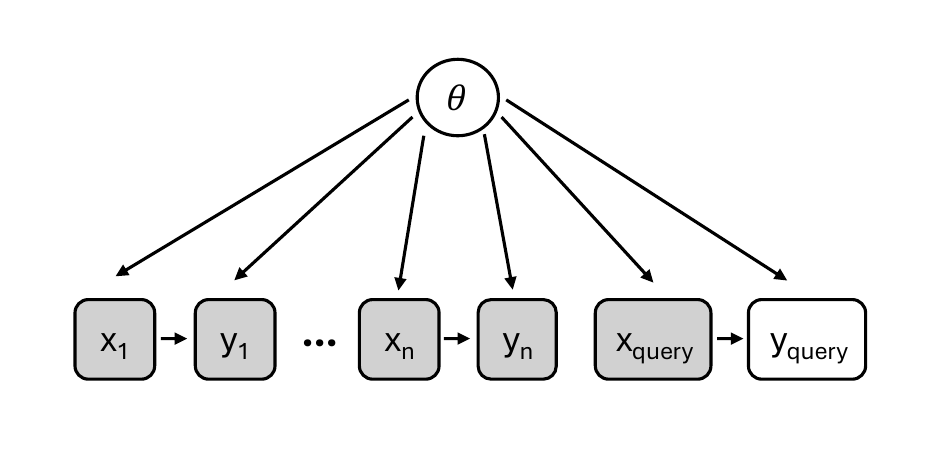}
    \caption{ICL as Bayesian inference}
    \label{fig:bayesian}
\end{figure}

\textbf{Bayesian perspective.} In-context learning can be viewed as approximate Bayesian inference, whether implicitly \cite{xie2022explanation} or explicitly \cite{garg2022transformers}. It has be shown that a Transformer directly fits the posterior predictive distribution (PPD), with a hypothesis latent concept $\phi$ which could parametrize the transitions of a Hidden Markov model \cite{baum1966statistical}, or a space of structural causal models. In this framework, the prior defines a space of hypotheses $\Phi$ on the relationship of a set of inputs $x$ to the output labels $y$. Each hypothesis $\phi \in \Phi$ can be seen as a mechanism that generates a data distribution from which one can draw context data with observed labels ${x_1,y_1,\ldots,x_n,y_n}$, denoted as ${c_{txt}}$, and an unlabeled query sample $x_{q}$, as illustrated in Figure \ref{fig:bayesian}.

The PPD specifies the distribution of the label $p(y_{q} | x_{q}, c_{txt})$ and can be obtained by integration over the space of hypotheses $\Phi$, where the weight of a hypothesis $\phi \in \Phi$ is determined by the posterior probability $p(\phi | c_{txt})$ given the context data. Considering the unnormalized posterior, which corresponds to the prior probability $p(\phi)$ and the likelihood $p(c_{txt} | \phi)$ of the training data given $\phi$, we have:

\begin{equation}
p(y_{q}|c_{txt},x_{q}) = \int_{\Phi} p(y_{q}| x_{q}, \phi) p(c_{txt} |\phi) p(\phi) d\phi
\end{equation}

\textbf{Frequentistic perspective.} Despite contexts being sampled from a different distribution than the pretraining distribution, the asymptotic prediction error of in-context learning is optimal when the signal about the latent concept in each context example is larger than the error due to the distribution mismatch. As a more informative and representative input space is provided, the learning error decreases \cite{xie2022explanation}. In this regard, one can also take a frequentistic perspective \cite{nagler2023statistical}. From a predictor's variance standpoint, being a pre-tuned, but untrained predictor with many hyperparameters and multi-head attention, a LTM has extremely high sensitivity to individual training samples, which translates in an increased ability to choose submodels and vanishing variance. From a bias standpoint, hyperparameters are pre-tuned to be optimal for a set of tasks defined by the prior. If this prior has large enough support and does not concentrate too much away from the true hypothesis, one can guarantee that the PPD converges to a close approximation of the true predictive distribution. Consequently, whether the LTM predictor can learn at inference time depends on its structural properties, with the optimal LTM approximation being characterized by a Kullback-Leibler criterion \cite{nagler2023statistical}. To allow for accurate approximation of the conditional class probabilities, one needs sufficiently complex LTM models and prior. Practically, a LTM is trained on simulated data sets. The larger these data sets are, the more complex the PPD is approximated. The training set size can therefore be understood as a regularizer on the expected complexity of the network. Thus, while LTM implementations ensure only vanishing predictor's variance, the bias vanishes only if the provided data is somewhat localized around the test feature.

\section{Context optimization}

Building on the theoretical insights into how LTMs can perform in-context stream learning, we now turn to the challenge of regulating their behavior in non-stationary environments. Specifically, the objective of context optimization is to minimize bias in model predictions when faced with new data that drifts over time. This problem occurs in situations where the LTM \( f_\theta \), is applied on-the-fly to a continuous stream of real-world data. The data stream is assumed to be partitioned sequentially into non-overlapping segments, called concepts, such as \( C_i \) and \( C_{i+1} \), each corresponding to distinct joint data distributions, \( p_i(x, y) \) and \( p_{i+1}(x, y) \), respectively. Within each concept, the data distribution is assumed static, however, as the data stream progresses, drift between consecutive concepts can introduce discrepancies between past and future data. For instance, if concept \( C_i \) has a skewed label distribution \( p_i(y) \), and the following concept \( C_{i+1} \) has a uniform label distribution \( p_{i+1}(y) \), then the predictive model must adapt to account for this shift in label distribution \cite{nejjar2024context}. The goal is to select an optimal localized context $l$, denoted by \( (c_{txt}, x_q) \), which helps \( f_\theta \) approximate the correct output \( y_q \) even for cases in the tail regions of these drifting distributions. In this case, the model’s expected prediction error for a new sample \( (x_q, y_q) \) is defined by:
\begin{equation}
\begin{split}
& \mathbb{E}_l[(f_\theta(x_q \mid c_{txt}) - \mathbb{E}[y_q \mid x_q])^2] \\
&= \mathbb{E}_l[( (f_\theta(x_q \mid c_{txt}) - \mathbb{E}_l[f_\theta(x_q \mid c_{txt})]) \\
&+ (\mathbb{E}_l[f_\theta(x_q \mid c_{txt})] - \mathbb{E}[y_q \mid x_q]) )^2] \\
&= \text{Var}[f_\theta(x_q \mid c_{txt})] + \left( \text{Bias}^2[f_\theta(x_q \mid c_{txt})] \right) + \sigma^2
\end{split}
\end{equation}

\textbf{LTM's variance decreases with more in-context examples.} Assume \( f_\theta \) to be \( c \)-Lipschitz continuous with constant \( c = (c_1, \ldots, c_n) \) where each \( c_i \) scales as \( \delta i^{-\alpha} \), \( \delta \) is a positive constant, and \( \alpha > 0.5 \). Then, for any two independently sampled contexts \( (c_{txt}, x_q) \) and \((c_{txt}', x_q) \) from an unbounded stream \( D \), the following inequality $|f_\theta(x_q \mid (c_{txt})) - f_\theta(x_q \mid (c_{txt}'))| \leq \sum_{i=1}^n c_i \mathbf{1}_{\{x_i \neq x'_i\}}$ holds \cite{garg2022transformers}. Applying McDiarmid’s Inequality \cite{mcdiarmid1989method}, for any \( t > 0 \), the tail probability bound is:
\begin{equation}
Pr(|f_\theta(x_q \mid c_{txt}) - \mathbb{E}_l[f_\theta(x_q \mid c_{txt})]| \geq t) \leq 2 \exp\left(-\frac{2t^2}{|c|^2_2}\right)
\end{equation}

where \( |c|^2_2 = \sum_{i=1}^{\infty}(\delta i^{-\alpha})^2 \), converging due to \( \alpha > 0.5 \). By the Borel-Cantelli lemma, \( f_\theta(x_q \mid c_{txt}) \) converges almost surely to the expected prediction \( \mathbb{E_s}[f_\theta(x_q \mid c_{txt})] \) as \( n \rightarrow \infty \) \cite{nejjar2024context}. This setup implies that as more samples are added to the prompt, the model’s sensitivity to small input changes diminishes, reducing error volatility.

\textbf{LTM's biases inherited from in-context examples.} Assume \( f_\theta \) mimics gradient descent steps on examples \( (c_{txt}, x_q) \), the model’s average prediction over many samples \( \mathbb{E}[f_\theta(x_q \mid c_{txt})] \) almost surely converges to the true conditional expectation \( \mathbb{E}[y_q \mid x_q] \) as \( n \to \infty \), provided there is no drift in the data distribution \cite{von2022transformers}. However, if there is shift in the distribution, \( (c_{txt}, x_q) \) may become misaligned with the current distribution, leading to biased predictions that reflect outdated patterns. This creates a tension between retaining examples from old concepts \(\{C_j\}_{j=1}^{i-1} \) just for the global representation of all classes \(y \in \mathcal{Y} \), and adapting to transient local patterns representing the most up to date concept \( C_i \), at the cost of not representing all classes. 

\begin{figure}[ht]
    \centering
    \includegraphics[width=0.40\textwidth]{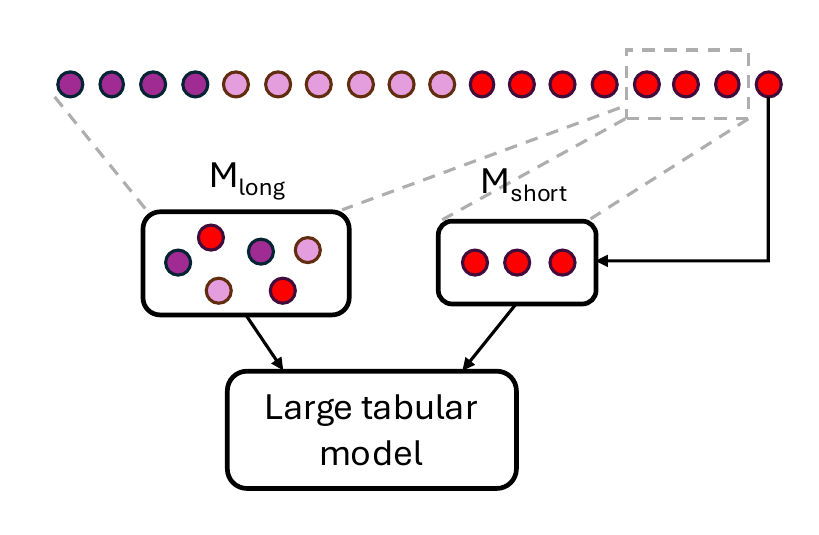}
    \caption{Short- and long-term memories}
    \label{fig:naive}
\end{figure}

\textbf{Dual-memory.} To keep representative samples of reoccurring classes, while gradually incorporating more data from recent classes, one can create a dual-memory FIFO mechanism that combines long-term inter-class balance with short-term intra-class responsiveness \( c_{\text{txt}} = S_t \cup L_t \), as depicted in Figure \ref{fig:naive}. To maintain stable inter-class boundaries over time, a long-term memory \( L_t \subset D_{<t} \) of size \( M_{\text{long}} \) retains temporally older samples that ensure a balanced distribution across all known classes \(\mathcal{Y} \). This memory helps preserve representation from rare or fading classes, especially when their frequency \( p_i(y) \ll 1 \). Over time, minority classes may evolve from outlier regions toward denser areas, or even swap roles with majority classes as the stream progresses. The long-term memory ensures that the context remains representative of the most up to date global distributional structure over extended periods. Complementarily, a short-term memory \( S_t \subset D_{t-M_{\text{short}}:t} \) captures the most recent samples from the stream. This memory focuses on preserving intra-class variability and recent sub-distributions \( p_t(x \mid y) \). Within a single class \(y \in \mathcal{Y} \), examples may cluster into temporally local sub-concepts, separated in input space due to drift. These sub-clusters may expand, shift, or disappear with time. The short-term buffer tracks these fine-grained dynamics, enabling the model to adapt to class recurrence, sub-concept emergence, and local outlier patterns.

\begin{table*}[!ht]
    \centering
    \caption{Prequential accuracy (*: no statistical difference), NC = No-change, MC = Majority-Class}
    \label{tab:results}
    \begin{tabular}{p{1.3cm}|p{1.6cm}p{1.6cm}p{1.7cm}p{1.8cm}p{1.6cm}p{1.5cm}p{1.5cm}p{0.7cm}p{0.7cm}}
        \toprule
        Dataset & LTM & ARF (n=90) & BOLE (n=50) & LevBag (n=45) & SRP (n=80) & EFDT & VFDT & NC & MC \\ \midrule
        NOAA & \textbf{81.29 ± 0.26} & 78.36 ± 0.20 & 73.45 ± 0.51 & 75.51 ± 0.23 & 78.35 ± 0.18 & 72.34 ± 0.43 & 70.86 ± 1.56 & 68.03 & 68.62 \\
        METER & \textbf{73.62 ± 0.28} & 68.02 ± 0.36 & 67.28 ± 1.54 & 63.17 ± 5.78 & 69.69 ± 1.81 & 57.94 ± 2.21 & 52.07 ± 2.63 & 1.08 & 8.22 \\
        ELEC & \textbf{91.82 ± 0.18} & 87.89 ± 0.22 & 90.17 ± 0.98\textsuperscript{\textcolor{red}{*}} & 85.86 ± 0.16 & 88.70 ± 0.57 & 77.96 ± 1.28 & 77.87 ± 1.29 & 85.30 & 57.41 \\
        RIALTO & \textbf{79.03 ± 1.06} & 67.95 ± 0.36 & 47.27 ± 1.54 & 60.99 ± 5.78 & 75.12 ± 1.81 & 55.79 ± 2.21 & 28.68 ± 2.63 & 00.00 & 10.00 \\
        POSTURE & \textbf{59.72 ± 0.18} & 59.14 ± 0.19\textsuperscript{\textcolor{red}{*}} & 48.28 ± 1.09 & 55.44 ± 0.69 & 58.14 ± 0.55 & 48.91 ± 0.18 & 52.99 ± 1.05 & 20.07 & 27.29 \\
        COVER & \textbf{95.53 ± 0.23} & 91.75 ± 0.14 & 93.33 ± 0.59\textsuperscript{\textcolor{red}{*}} & 85.26 ± 0.24 & 94.54 ± 0.20\textsuperscript{\textcolor{red}{*}} & 82.38 ± 1.08 & 76.98 ± 2.37 & 95.06 & 48.76 \\
        POKER & \textbf{97.51 ± 0.31} & 89.17 ± 0.43 & 80.01 ± 1.54 & 88.52 ± 3.12 & 87.58 ± 0.88 & 77.38 ± 4.10 & 77.43 ± 7.02 & 74.54 & 50.11 \\
        \midrule
        Average & \textbf{82.65} & 77.47 & 71.40 & 73.54 & 78.87 & 67.53 & 62.41 & 49.15 & 38.63 \\
        \midrule
        Rank & \textbf{1.00} & 2.86 & 4.57 & 4.29 & 2.71 & 6.00 & 6.57 & 7.43 & 8.43 \\
        \bottomrule
    \end{tabular}
\end{table*}

\textbf{FIFO algorithm.} The pseudocode is given in Algorithm \ref{alg:ltm}. It begins by initializing the short-term memory $S$ and long-term memory $L$, along with a class count dictionary $C[y]$ for each class $y$. The size of the long-term memory is computed as $M_{\text{long}} = M - M_{\text{short}}$. After a warming period $T_{\text{warm}}$, the algorithm enters the main loop, processing each new sample $(x_t, y_t)$ from the data stream $D_{\text{stream}}$. For each new sample, the algorithm first adds it to the short-term memory $S$. If the short-term memory exceeds its maximum size $M_{\text{short}}$, the oldest sample is removed and moved to the long-term memory $L$. The class count for the corresponding class of the removed sample is updated in the dictionary $C$. Once the long-term memory exceeds its capacity $M_{\text{long}}$, the algorithm identifies the most overrepresented class, $y_{\max}$, in long-term memory $L$. The oldest sample from this overrepresented class is then removed to maintain balance in the memory, ensuring that the system does not retain too many samples from any single class.

\begin{figure}[ht]
    \centering
    \begin{algorithm}[H]
        \caption{Naive context optimization}
        \label{alg:ltm}
        \begin{algorithmic}[1]  
            \STATE \textbf{Input:} $D_{\text{stream}}$, $M_{\text{short}}$, $M_{\text{long}}$, $T_{\text{warm}}$, $LTM$
            \STATE $S, L \gets \emptyset$
            \STATE $C[y] \gets 0, \forall y \in \mathcal{Y}$
            \STATE $t_{\text{warm}} \gets 0$
            
            \FOR{each new sample $(x_t, y_t) \in D$}
                \IF{$t_{\text{warm}} > T_{\text{warm}}$}
                    \STATE $\hat{y} \gets \text{learn in-context}(LTM, S \cup L)$
                    \STATE $A_T \gets \frac{1}{T} \sum_{t=1}^{T} \mathbb{1} (\hat{y}_t = y_t)$
                \ENDIF
                
                \STATE $S \gets S \cup \{(x_t, y_t)\}$
                
                \IF{$|S| > M_{\text{short}}$}
                    \STATE $(x_{\text{old}}, y_{\text{old}}) \gets$ oldest sample in $S$
                    \STATE $S \gets S \setminus \{(x_{\text{old}}, y_{\text{old}})\}$
                    \STATE $L \gets L \cup \{(x_{\text{old}}, y_{\text{old}})\}$
                    \STATE $C[y_{\text{old}}] \gets C[y_{\text{old}}] + 1$
                \ENDIF
                
                \IF{$|L| > M_{\text{long}}$}
                    \STATE $y_{\max} \gets \arg\max_{y \in \mathcal{Y}} C[y]$
                    \STATE $(x_{\text{old}}, y_{\text{old}}) \gets$ oldest sample in $\{ (x, y) \in L \mid y = y_{\max} \}$
                    \STATE $L \gets L \setminus \{(x_{\text{old}}, y_{\text{old}})\}$
                    \STATE $C[y_{\text{old}}] \gets C[y_{\text{old}}] - 1$
                \ENDIF
            \ENDFOR
        \end{algorithmic}
    \end{algorithm}
\end{figure}

\section{Experimental study}
\label{sec:experiments}

This section evaluates the performance of the proposed context optimization of a large tabular model (LTM) against state-of-art algorithms for data stream mining, namely: Adaptive Random Forest (ARF) \cite{gomes2017adaptive}, Streaming Random Patches (SRP) \cite{gomes2019streaming}, Boosting-like Online Learning Ensemble (BOLE) \cite{de2016boosting}, Leverage Bagging (LevBag) \cite{bifet2010leveraging}, Extremely Fast Decision Tree (EFDT) \cite{manapragada2018extremely}, and Very Fast Decision Tree (VFDT) \cite{domingos2000mining}. These algorithms were implemented within the Massive Online Analysis (MOA) framework\footnote{https://github.com/Waikato/moa}, and tuned for each dataset using a grid search strategy, with ensembles of up to 90 components, grace periods set to 100, 400, and 1000, and tie-splitting thresholds configured to 0.01, 0.05, and 0.1 as the hyperparameter settings. The Hoeffding bound confidence level was fixed at $1 \times 10^{-7}$ across all experiments. Adaptive Naive Bayes prediction was used at the leaf level for all models. For the LTM, we use the publicly available first version of the transformer-based TabPFN\footnote{https://github.com/PriorLabs/TabPFN}. A key practical advantage of LTMs is their minimal reliance on hyperparameter tuning. In TabPFN, the primary source of variability arises from the ensemble prediction mechanism, in which feature orderings and scalings are randomly permuted to account for the permutation invariance of tabular columns. In our experiments, 4 permutations were used. Increasing the number of permutations reduces stochasticity and yields a closer approximation to the Bayesian posterior, typically resulting in monotonic performance improvements. This behavior contrasts with tree-based ensembles, where adding models introduces explicit bias–variance–diversity trade-offs. The proposed dual-memory mechanism also introduces very few parameters. For context optimization, $M$ was set to 600, 800, and 1000 instances, and $M_{\text{short}}$ to 0.65, 0.75, 0.85 instances. Moreover, an initial warming period $T_{\text{warm}}$ of 100 instances was considered to collect samples into the memory, without making predictions.

\begin{figure*}[h]
    \centering
    \includegraphics[width=\textwidth]{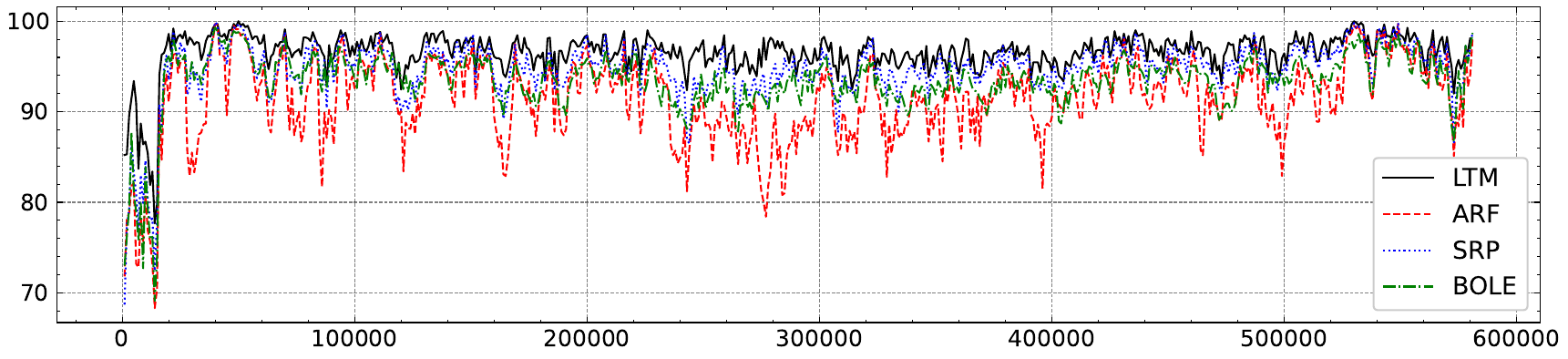}
    \caption{Prequential accuracy over time}
    \label{fig:results2}
\end{figure*}

\begin{figure}[ht]
    \centering
    \includegraphics[width=0.48\textwidth]{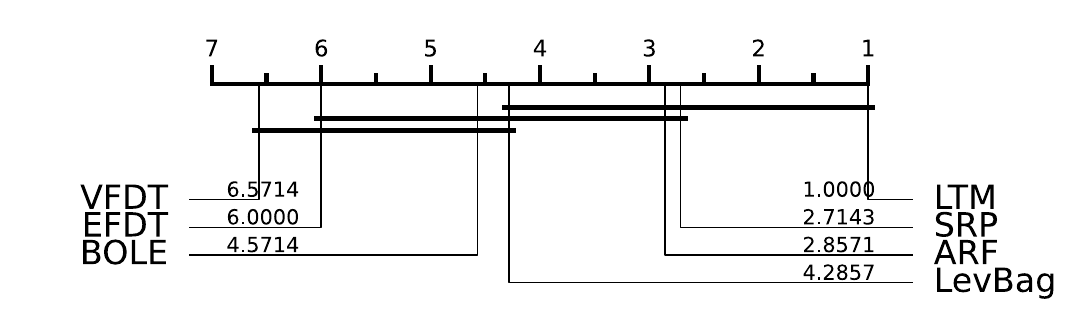}
    \caption{Nemeyi test at p-value = 5\%}
    \label{fig:cdd}
\end{figure}

In evaluating the application of any foundation model, one can consider both intrinsic and extrinsic evaluation metrics \cite{bommasani2022opportunities}. Extrinsic evaluation refers to measuring the performance of the adapted foundational model on a downstream task, whereas intrinsic evaluation refers to directly measuring the foundational model's quality. Since this work is concerned with the adaptability of LTMs to streams, only extrinsic performance evaluation is performed, using a prequential evaluation strategy, where each instance is first used to test then update the classifier in an online manner (instance by instance) \cite{gama2013evaluating}. The used data streams are from the USP data stream repository\footnote{https://sites.google.com/view/uspdsrepository}, encompassing both binary and multiclass classification tasks: NOAA Weather (18,159 instances), SmartMeter LADPU (22,950), Electricity (45,312), Rialto Bridge Timelapse (82,250), Posture (164,860), Forest CoverType (581,012) and PokerHand (829,201). Table~\ref{tab:results} and Figure~\ref{fig:cdd} compare LTM against all baseline algorithms. LTM achieves the highest overall performance, with an average accuracy of 82.65\% and a mean rank of 1. The strongest baselines, ARF and SRP, reach 77.47\% and 78.87\%, respectively. The largest improvements are observed on the POKER, METER, RIALTO, and NOAA datasets, where LTM exceeds the best alternative by 8.34 pp, 3.93 pp, 3.91 pp, and 2.93 pp, respectively. Two-tailed t-tests (p~$<$~0.05) confirm LTM’s superiority in all cases except against ARF on POSTURE, BOLE on ELEC/COVER, and SRP on COVER.

Grid searches over total memory sizes \{600, 800, 1000\} and short-term ratios \{0.65, 0.75, 0.85\} produce highly stable accuracies, with standard deviations up to only 0.31, with the exception of RIALTO. In contrast, several baselines exhibit substantial sensitivity to hyperparameters. Single-tree models such as EFDT and VFDT show standard deviations as high as 4.10 and 7.02, while ensemble methods such as SRP, LevBag, and BOLE can still fluctuate up to 1.81, 5.78, and 1.54, respectively. These results highlight LTM’s robustness and consistency across conditions. Table \ref{tab:ablation} reports the ablation study of each individual component of the dual-memory strategy. Incorporating the dual strategy consistently improves performance over the long-term component alone, with an average gain of +3.24 (range +2.33 to +4.59). These gains are attenuated in datasets with large number of classes and highly volatile class distributions, as reflected by the weak performance of the Majority-Class baseline. The dual strategy also improves over short-term memory alone (+0.31 on average), though gains are smaller and sometimes negative (ELEC, COVER). This is likely attributable to strong temporal dependencies, as suggested by the high accuracy of the No-Change classifier. Increasing context size from 600 to 1000 yields an average improvement of +0.75, statistically significant for most datasets.

\begin{table}[!ht]
\centering
\caption{Ablation study (*: no statistical difference)}
\label{tab:ablation}
\begin{tabular}{l|p{1.6cm}p{1.6cm}p{1.6cm}}
\toprule
Dataset & Dual - Long & Dual - Short & 1000 - 600 \\
\midrule
NOAA & +4.59 & +0.89 & +0.50\textsuperscript{\textcolor{red}{*}} \\
METER & +3.35 & +0.27\textsuperscript{\textcolor{red}{*}} & +0.98 \\
ELEC & +4.11 & -0.22\textsuperscript{\textcolor{red}{*}} & +0.89 \\
RIALTO & +3.45 & +0.91 & +1.54 \\
POSTURE & +2.33 & +0.24\textsuperscript{\textcolor{red}{*}} & +0.39\textsuperscript{\textcolor{red}{*}} \\
COVER & +2.48 & -0.12\textsuperscript{\textcolor{red}{*}} & +0.46\textsuperscript{\textcolor{red}{*}} \\
POKER & + 2.35 & +0.18\textsuperscript{\textcolor{red}{*}} & +0.47 \\
\midrule
Average & +3.24 & +0.31 & +0.75 \\
\bottomrule
\end{tabular}
\end{table}

Figure \ref{fig:results2} illustrates the evolution of accuracy over time for COVER, showing that LTM experiences significantly smaller performance drops at drift points. While these baselines re-stabilize quickly, LTM’s windowing approach excels in handling drifts, maintaining a strong focus on recent instances. In extended intervals of stable concepts, however, the baselines tend to converge toward LTM’s accuracy. To further substantiate the results, Table \ref{tab:otherresults} presents the reported accuracies from the original papers of the various baselines, including not only their own performance but also the results of other algorithms: Online Smooth Boosting (OB) \cite{chen2012online}, Online Accuracy Updated Ensemble (OAUE) \cite{brzezinski2014combining}, Dynamic Weighted Majority (DWM) \cite{kotler2003dynamic}, and Streaming Gradient Boosted Trees (SGBT) \cite{gunasekara2024gradient}. Notably, none report results outperforming LTM. 

\begin{table}[!ht]
    \centering
    \caption{Prequential accuracy reported in other benchmarks}
    \label{tab:otherresults}
    \begin{tabular}{c|cccccccc}
    \toprule
        \multirow{2}{*}{Model} & \multicolumn{2}{c}{SGBT$^\ast$/BOLE$^\dagger$} & \multicolumn{2}{c}{SRP (n=100)} & \multicolumn{2}{c}{ARF (n=100)} \\ 
        & ELEC & COVER & ELEC & COVER & ELEC & COVER \\ \midrule
        ARF & 90.62$^\ast$ & 94.72$^\ast$ & 88.71 & 94.69 & 89.67 & 94.97 \\ 
        LevBag  & 89.71$^\dagger$ & 88.13$^\dagger$ & 90.16 & 94.86 & 89.51 & 95.1 \\ 
        SRP  & 89.68$^\ast$ & 95.34$^\ast$ & 88.82 & 95.25 & 89.86 & 95.35 \\ 
        OB  & 89.51$^\ast$ & 92.69$^\ast$ & 85.25 & 90.33 & 89.52 & 92.7 \\ 
        OAUE & ~ & ~ & 88.28 & 90.17 & 87.41 & 92.86 \\ 
        DWM  & 88.52$^\dagger$ & 87$^\dagger$ & 87.76 & 88.29 & 87.76 & 88.52 \\ 
        SGBT  & 88.5$^\ast$ & 94.29$^\ast$ & ~ & ~ & ~ & ~ \\
        BOLE & 90.04$^\dagger$ & 90.16$^\dagger$ & ~ & ~ & ~ & ~ \\ \midrule
        LTM & \textbf{91.82} & \textbf{95.53} & \textbf{91.82} & \textbf{95.53} & \textbf{91.82} & \textbf{95.53} \\ \bottomrule
    \end{tabular}
\end{table}

While the performance of LTM is well-established, it is essential to evaluate its adherence to key stream mining requirements \cite{bifet2009adaptive,lourencco2025device}: (R1) processing one example at a time and inspecting it only once, (R2) using a limited amount of memory, (R3) operating within a limited time budget, and (R4) being ready to make predictions at any point. Regarding (R1), no strict constraint prevents an algorithm from temporarily storing examples internally. As long as these stored instances are eventually discarded to satisfy (R2), an algorithm remains compliant with this requirement. In terms of (R2), although LTMs may require substantial memory storage, their memory footprint remains constant, enabling them to handle data volumes far exceeding the available working memory. In this regard, memory limitations are a physical constraint that must be balanced with adherence to (R3), i.e., ensuring feasible runtime performance. The memory in LTMs can be divided into two categories: the one allocated for running statistics of long- and short-term memory and the memory used to store the foundational model. While data stream mining algorithms typically optimize memory efficiency by maintaining running statistics that directly inform predictions, LTMs inherently separate these two types of memory. For (R3), LTMs must scale efficiently to any number of examples, which is ensured by its linear runtime complexity concerning the number of processed instances. Furthermore, the computational cost per example is upper-bounded by the number of features, classes and instances within the given context. A critical requirement for real-time operation is that the LTM processes instances as fast as they arrive, as failure to do so leads to data loss. For reference, in MOA, an ARF with 100 hoeffding trees processes 100k samples with 30 features in 2.4~ms per instance. While LTMs may be slower, several strategies can mitigate this, depending on the architecture. For example, Table~\ref{fig:requirement} reports the processing time per sample when using TabPFN on a NVIDIA Tesla T4 GPU to predict ten instances in parallel in a single forward pass. Alternatively, a hypernetwork-based LTM \cite{bonet2024hyperfast} detaches the inference cost of the LTM from that of the prediction by the generated network, suggesting a different strategy. Classic stream management techniques can also be applied on an application-specific basis to ensure timely predictions. For example, instances can be offloaded to other learners or buffered for processing during idle periods. Finally, concerning (R4), while conventional stream learners continuously update their models based on evolving statistics, a LTM translates these statistics into a structured representation, incorporating a selected subset of in-context examples. 

\begin{table}[ht]
    \centering
    \caption{Processing time per instance}
    \label{fig:requirement}
    \begin{tabular}{c|ccc}
    \toprule
    \textbf{Name}  & \textbf{Classes} & \textbf{Features} & \textbf{Time (ms)} \\ \midrule
    NOAA     & 2              & 8            &   18.4    \\ 
    METER        & 10             & 96              &   81.4   \\ 
    ELEC     & 2              & 8                 &   16.9  \\ 
    RIALTO        & 10             & 27               &   28.7   \\ 
    POSTURE         & 11             & 3               &   20.3  \\ 
    COVER   & 7              & 54                 &  41.1  \\ 
    POKER       & 10             & 11              &     24.2   \\ \bottomrule
    \end{tabular}
\end{table}

\section{Conclusions}
\label{sec:discussion}

In the early stages of machine learning, the dominant belief was that the best way to solve a task was to meticulously design a specific dataset and tailor a model exclusively for that task. Over time, it became evident that for certain types of unstructured data, such as image recognition, a general-purpose model could be fine-tuned on specific datasets, yielding effective results. Today, we stand at a point where highly versatile models, such as general-purpose transformers, can be prompted to tackle a wide range of tasks without task-specific training. With this work, we embrace this evolving paradigm. So far, we have identified LTMs as a promising candidate for streaming solutions, warranting further research. The key challenge is not merely the size of the model or processing speed but rather the dynamic interplay between data arrival, training, recovery, and inference. As technology advances, models will naturally become smaller, faster, and more affordable. In the meantime, efforts should focus on designing algorithms that align more closely with the core principles of data stream mining: processing data in its natural order, handling labels that may be correct at the time of request but outdated or even incorrect by the time they become available, minimizing interference among concepts, and so on.

Building on this foundation, several promising research directions emerge within the main goal of context optimization in selecting data with properties similar to the desired target distribution. For instance, distribution matching does not need to be constrained to a uniform context across all query points. Instead, a more flexible approach involves storing a larger pool of encoded knowledge that can be dynamically grouped into adaptive local contexts, tailored to each specific query \cite{bonet2024hyperfast}. This retrieval-based strategy allows to go beyond the proposed temporal assumption, where recent points are likely to be of the same concept. By first applying a distribution diversification method to prioritize heterogeneity in a sample and remove redundancies, distribution matching methods can operate within a representation space that allows for similarity to be measured across data points. Thus allowing to preserve valuable information that could be useful in the future and adopt new inductive biases on how to match data distributions, such as: the smoothness assumption, where if two points reside in a high-density region are close, then so should be their corresponding outputs; the cluster assumption, where if points are in the same cluster, they are likely to be of the same concept; the manifold assumption, where high-dimensional data lies in a low-dimensional manifold. For this purpose, online synopsis techniques, such as histograms, wavelets, sketches can be used to incrementally capture the internal structure of the classes \cite{silva2013data}, and followed by an offline mining process performed on the stored micro-cluster synopsis whenever a LTM sends a request. Further visionary perspectives on these challenges and opportunities were discussed at the \textit{Streaming Continual Learning AAAI Bridge 2026} \cite{lourencco2025bridging}. The authors also plan to pursue several of these directions and welcome feedback and collaboration.

\begin{acks}
Work funded by Portugal 2030 and the national Foundation for Science and Technology under Ph.D. scholarship PRT/BD/154713/2023, project FOXPM (COMPETE2030-FEDER-00923300 – 14731), and project doi.org/10.54499/UIDP/00760/2020.
\end{acks}

\bibliographystyle{ACM-Reference-Format}
\bibliography{references}

\end{document}